%% file: main.tex
\documentclass[runningheads]{llncs}

\usepackage{eccv}

\usepackage{eccvabbrv}

\usepackage{graphicx}
\usepackage{booktabs}

\usepackage{mwe}
\usepackage{subcaption}
\usepackage{calc}

\usepackage[accsupp]{axessibility}  

\usepackage{hyperref}

\usepackage{orcidlink}

\usepackage{tabularx}
\newcolumntype{Y}{>{\centering\arraybackslash}X}
\newcolumntype{C}[1]{>{\centering}p{#1}}
\usepackage{caption}
\usepackage{subcaption}

\begin{document}

\title{SWinGS: Sliding Windows for\\Dynamic 3D Gaussian Splatting} 

\titlerunning{SWinGS}

\author{Richard Shaw\inst{1} \and
Michal Nazarczuk\inst{1} \and
Jifei Song\inst{1} \and
Arthur Moreau\inst{1} \and
Sibi Catley-Chandar\inst{1,2} \and
Helisa Dhamo\inst{1} \and
Eduardo P\'erez-Pellitero\inst{1}}

\authorrunning{R.~Shaw et al.}

\institute{Huawei Noah's Ark Lab \and
Queen Mary University Of London}

\maketitle

\begin{figure}[h]
    \centering
    \begin{subfigure}[b]{0.61\textwidth}
         \centering
            \includegraphics[trim={0pt 45pt 900pt 0pt},clip,width=\linewidth]{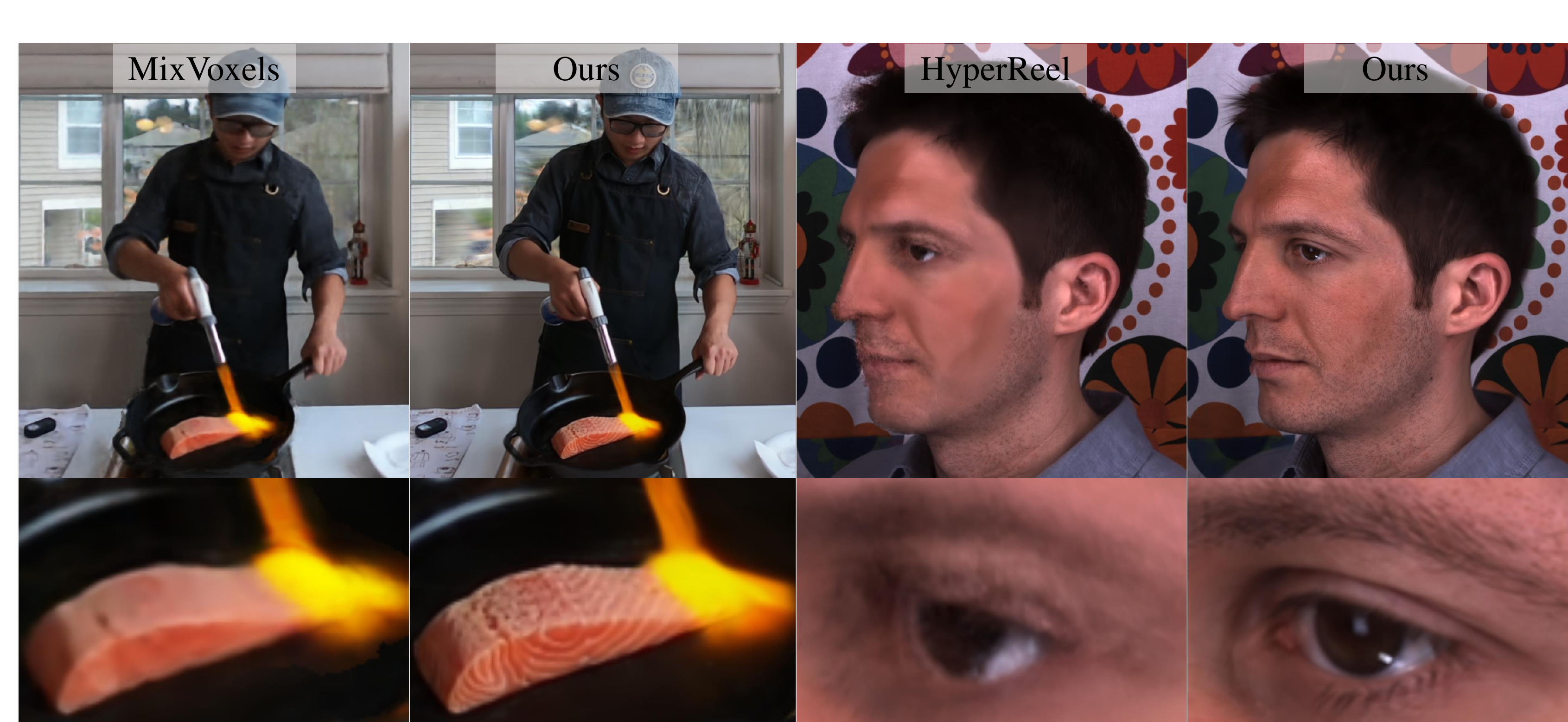}
         \label{fig:viewer_suba}
    \end{subfigure}
    \hfill
    \begin{subfigure}[b]{0.38\textwidth}
         \centering
         \includegraphics[trim={1650pt 86 0pt 90pt},clip,width=0.99\linewidth]{images/SWAGS_banner.pdf}
         \label{fig:viewer_subb}
     \end{subfigure}
    \caption{\textbf{Left:} SWinGS achieves sharper dynamic 3D scene reconstruction in part thanks to a sliding window canonical space that reduces the complexity of the 3D motion estimation. 
    \textbf{Right:} Our dynamic real-time viewer allows users to explore the scene.}
    \label{fig:viewer}
\end{figure}

\begin{abstract}
Novel view synthesis has shown rapid progress recently, with methods capable of producing increasingly photorealistic results. 3D Gaussian Splatting has emerged as a promising method, producing high-quality renderings of scenes and enabling interactive viewing at real-time frame rates. However, it is limited to static scenes. In this work, we extend 3D Gaussian Splatting to reconstruct dynamic scenes. We model a scene's dynamics using dynamic MLPs, learning deformations from temporally-local canonical representations to per-frame 3D Gaussians. To disentangle static and dynamic regions, tuneable parameters weigh each Gaussian's respective MLP parameters, improving the dynamics modelling of imbalanced scenes. We introduce a sliding window training strategy that partitions the sequence into smaller manageable windows to handle arbitrary length scenes while maintaining high rendering quality. We propose an adaptive sampling strategy to determine appropriate window size hyperparameters based on the scene's motion, balancing training overhead with visual quality. Training a separate dynamic 3D Gaussian model for each sliding window allows the canonical representation to change, enabling the reconstruction of scenes with significant geometric changes. Temporal consistency is enforced using a fine-tuning step with self-supervising consistency loss on randomly sampled novel views. As a result, our method produces high-quality renderings of general dynamic scenes with competitive quantitative performance, which can be viewed in real-time in our dynamic interactive viewer.
\end{abstract}

\input{sec/1_intro}
\input{sec/2_rw}
\input{sec/3_method}

\input{sec/4_results}

\input{sec/5_conclusion}

%
%
\bibliographystyle{splncs04}
\bibliography{main}
\end{document}

%% file: sec/1_intro.tex

\section{Introduction}
\label{sec:intro}

Photorealistic rendering and generally 3-dimensional (3D) imaging have received significant attention in recent years, especially since the seminal work of Neural Radiance Fields (NeRF)~\cite{mildenhall2020}. This is in part thanks to its impressive novel view synthesis results, but also due to its appealing ease of use when coupled with \textit{off-the-shelf} structure-from-motion camera pose estimation~\cite{Schnberger2016StructurefromMotionR}. NeRF's key insight is a fully differentiable volumetric rendering pipeline paired with learnable implicit functions that model a view-dependent 3D radiance field. Dense coverage of posed images of the scene provides then direct photometric supervision.

The original formulation of NeRF and most follow-ups~\cite{mildenhall2020, TensoRFChen2022ECCV, mueller2022instant, Barron2021MipNeRFAM, Niemeyer2021RegNeRFRN} assume static scenes and thus a fixed radiance field. Some have explored new paradigms enabling dynamic reconstruction for radiance fields, including D-NeRF~\cite{pumarola2020} and Nerfies~\cite{park2021a}, optimising an additional continuous volumetric deformation field that warps each observed point into a canonical NeRF. Such an approach has been popular~\cite{du2021, fang2022, li2020, liu2023, lombardi2019, park2021b}, and has also been used for dynamic human reconstruction~\cite{peng2021animatable, Zhao_2022_CVPR}. However, learning 3D deformation fields is inherently challenging, especially for large motions, with increased computational expense in training and inference. Moreover, approaches that share a canonical space among all frames struggle to maintain reconstruction quality for long sequences, obtaining overly blurred results due to inaccurate deformations and limited representational capacity. Other methods avoid maintaining a canonical representation and use explicit per-frame representations of the dynamic scene. Examples are tri-plane extensions to 4D $(x,y,z,t)$~\cite{cao2023, fridovich2023, shao2023, turki2023} with plane decompositions~\cite{TensoRFChen2022ECCV} to keep memory footprint under control. These approaches can suffer from a lack of temporal consistency, especially as they generally are agnostic about the motion of the scene. Grid-based methods that share a representation~\cite{peng2023}, also can suffer from degradation due to lack of representational capacity in long sequences.

This paper proposes a new method that addresses open problems of the state-of-the-art (SoTA) - see Fig.\,\ref{fig:viewer}. Our method overview is shown in Fig.\,\ref{fig:overview}.
Firstly, we build upon 3D Gaussian Splatting (3DGS)~\cite{kerbl2023} and adapt the 3D Gaussians to be dynamic by allowing them to move. Our representation is thus explicit and avoids expensive raymarching via fast rasterization.
Secondly, we introduce a novel paradigm for dynamic neural rendering with temporally-local canonical spaces defined in a sliding window fashion. Each window's length is adaptively defined following the amount of scene motion to maintain high-quality reconstruction. By limiting the scope of each canonical space, we can accurately track 3D displacements (\ie they are generally smaller displacements) and prevent intra-window flickering. 
Thirdly, we introduce tuneable MLPs (MLP with several sets of weights governed by per-3D Gaussian blending weights) to estimate displacements. This tackles scenes with static vs dynamic imbalance. By learning different ``modes'' of motion estimation, we can separate smoothly between static and dynamic regions with virtually no additional computational cost nor any handcrafted heuristics.
Lastly, temporal consistency loss computed on overlapping frames of neighbouring windows ensures consistency between windows, \ie avoids inter-window flickering. In summary, our main contributions are:
\begin{enumerate}
    \item An adaptive sliding window approach that enables the reconstruction of arbitrary length sequences whilst maintaining high rendering quality.

    \item Temporally-local dynamic MLPs that model scene dynamics by learning deformation fields from per-window canonical 3D Gaussians to each frame.

    \item Learnable MLP tuning parameters  tackle scene imbalance by learning different \textit{motion modes}; disentangling static canonical and dynamic 3D Gaussians.
    
    \item A fine-tuning stage ensures temporal consistency throughout the sequence.
\end{enumerate}

%% file: sec/2_rw.tex
\section{Related work}

\textbf{Non-NeRF dynamic reconstruction:}
A number of approaches prior to the emergence of NeRF~\cite{mildenhall2020} tackled dynamic 
 scene reconstruction. Such methods typically relied on dense camera coverage for point tracking and reprojection \cite{joo2014}, or the presence of additional measurements, \ie depth \cite{newcombe2015, slavcheva2017}. Alternatively, some were curated towards specific domains, \eg car reconstruction in driving scenarios \cite{barsan2018, luiten2020}. Several recent works \cite{bansal2020, li2023, peng2023} followed the idea of Image-Based Rendering \cite{wang2021} (direct reconstruction from neighbouring views). Others \cite{lin2021, xing2022} utilize multiplane images \cite{zhou2018, tanay2024} with an additional temporal component.

\noindent
\textbf{NeRF-based reconstruction:} NeRF~\cite{mildenhall2020} has achieved great success for reconstructing static scenes with many works extending it to dynamic inputs. Several methods~\cite{bansal2023, xian2021} model separate representations per time-step, disregarding the temporal component of the input. D-NeRF~\cite{pumarola2020} reconstruct the scene in a canonical representation and model temporal variations with a deformation field. This idea was developed upon by many works \cite{du2021, fang2022, li2020, liu2023, lombardi2019, park2021b}. StreamRF~\cite{li2022b} and NeRFPlayer~\cite{song2023} use time-aware MLPs and a compact 3D grid at each time-step for 4D field representation, reducing memory cost for longer videos. Other approaches~\cite{cao2023, fridovich2023, shao2023, turki2023} represent dynamic scenes with a space-time grid, with grid decomposition to increase efficiency. DyNeRF~\cite{li2022a} represents dynamic scenes by extending NeRF with an additional time-variant latent code. HyperReel~\cite{attal2023} uses an efficient sampling network, modeling the scene around keyframes. MixVoxels~\cite{wang2023} represents dynamic and static components with separately processed voxels. Several methods \cite{isik2023, li2022c, weng2022} rely on an underlying template mesh (\eg human). Some methods \cite{catleychandar2024roguenerf, rong2022bvs, zhou2023nerflix} aim to improve NeRF quality post rendering.

\noindent
\textbf{3D Gaussian Splatting} Much of the development of neural rendering has focused on accelerating inference \cite{hedman2021, reiser2021, peng2023, wang2022, yu2021}. Recently, 3D Gaussian Splatting~\cite{kerbl2023} made strides by modelling the scene with 3D Gaussians, which, when combined with tile-based differentiable rasterization, achieves very fast rendering, yet preserves high-quality reconstruction. Luiten~\etal~\cite{luiten2024} extend this to dynamic scenes with shared 3D Gaussians that are optimised frame-by-frame. However, their focus is more on tracking 3D Gaussian trajectories rather than maximizing final rendering quality. Concurrent methods to ours that extend 3D Gaussian Splatting to focus on general dynamic scenes include~\cite{yang2023gs4d, yang2023deformable3dgs, huang2023sc, lin2023gaussian,  li2023spacetime, sun20243dgstream} amongst others, while other works have focused specifically on dynamic human reconstruction~\cite{moreau2024human, li2024animatablegaussians, kocabas2024hugs, hu2024gaussianavatar, Pang_2024_CVPR, qian20233dgsavatar, Wu_2024_CVPR} and facial animation~\cite{headgas, Qian2023GaussianAvatarsPH, xiang2024flashavatar, saito2024rgca, xu2023gaussianheadavatar}.

%% file: sec/3_method.tex
\section{Method}
\label{sec:method}

\begin{figure}[t]
\centering
  \includegraphics[width=0.95\textwidth]{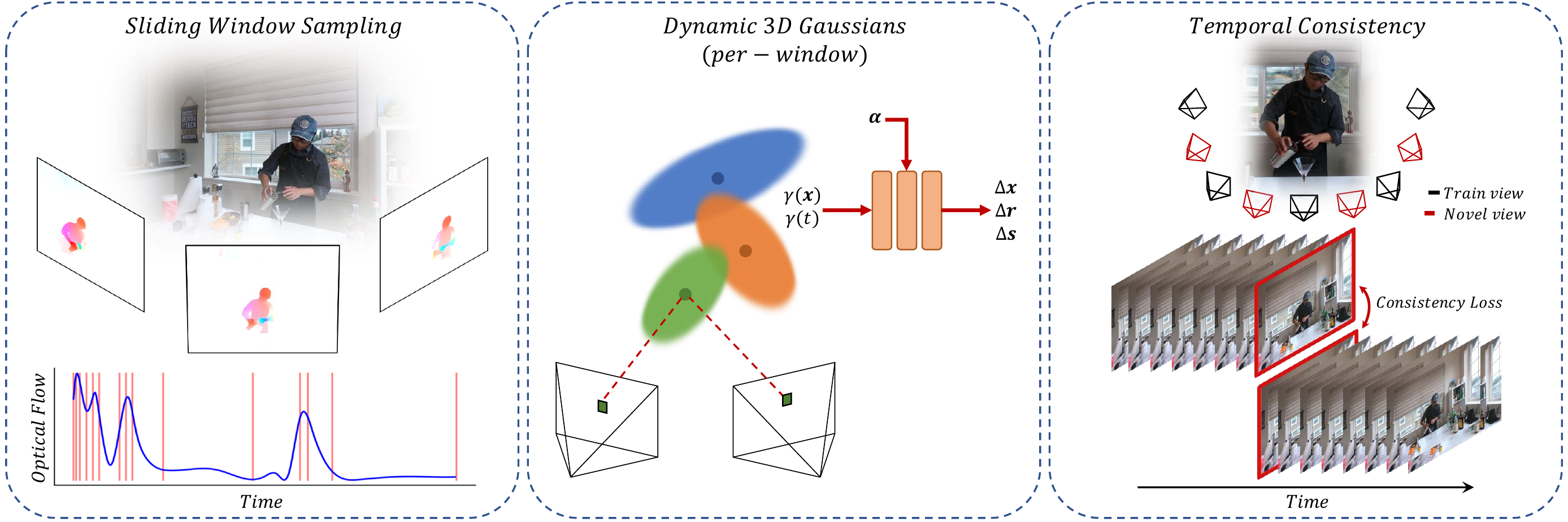}
  \caption{\textbf{Method.} First, the sequence is partitioned into sliding windows based on optical flow. Second, a dynamic 3DGS model is trained per window, where tunable MLPs model the deformations. Blending parameters $\boldsymbol{\alpha}$ weigh the MLP's parameters to focus on dynamic parts. Finally, each model is fine-tuned, enforcing inter-window temporal consistency with consistency loss on sampled views for overlapping frames.
  }
  \label{fig:overview}
\end{figure}

\subsection{Overview} We present our method to reconstruct and render novel views of general dynamic scenes from multiple calibrated time-synchronized cameras. An overview of our method is shown in Fig.~\ref{fig:overview}. We build upon 3D Gaussian Splatting~\cite{kerbl2023} for novel view synthesis of static scenes, which we extend to scenes containing motion.

Our method can be separated into three main steps. First, given a dynamic sequence, we split the sequence into separate shorter sliding windows of frames for concurrent processing. We adaptively sample windows of varying lengths depending on the amount of motion in the sequence. Each sliding window contains an overlapping frame between adjacent windows, enabling temporal consistency to be enforced throughout the sequence at a subsequent training stage. Partitioning the sequence into smaller windows enables us to deal with sequences of arbitrary length while maintaining high render quality.

Second, we train separate dynamic 3DGS models for each sliding window in turn. We extend the static 3DGS method to model the dynamics by introducing a tuneable MLP~\cite{Maggioni2023TunableCW}. The MLP learns the deformation field from a canonical set of 3D Gaussians for each frame in a window. Thus each window comprises an independent temporally-local canonical 3D Gaussian representation and deformation field. This enables us to handle significant geometric changes and/or if new objects appear throughout the sequence, which can be challenging to model with a single representation. A tuneable MLP weighting parameter $\boldsymbol{\alpha}$ is learned for each 3D Gaussian to enable the MLP to focus on modelling the dynamic parts of the scene, with the static parts encapsulated by the canonical representation.  

 Third, once a dynamic 3DGS model is trained for each window in the sequence, we apply a fine-tuning step to enforce temporal consistency throughout the sequence. We fine-tune each 3DGS model sequentially and employ a self-supervising temporal consistency loss on the overlapping frame renders between neighbouring windows. This encourages the model of the current window to produce similar renderings to the previous window. The result is a set of per-frame Gaussian Splatting models enabling high-quality novel view renderings of dynamic scenes with real-time interactive viewing capability. Our approach enables us to overcome the limitations of training with long sequences and to handle complex motions without exhibiting distracting temporal flickering.

\subsection{Preliminary: 3D Gaussian Splatting}

Our method is built upon 3D Gaussian Splatting (3DGS)~\cite{kerbl2023}. 3DGS uses a 3D Gaussian representation to model scenes as they are differentiable and can be projected to 2D splats, enabling fast tile-based rasterization. The 3D Gaussians are defined by 3D covariance matrix $\Sigma$ in world space centered at the mean $\mu$:

\begin{equation}
    G(x) = 
    e^{-\frac{1}{2}(x)^T \Sigma^{-1} (x)}.
\end{equation}

Given a scaling matrix $S$ and rotation matrix $R$, the corresponding covariance matrix $\Sigma$ of a 3D Gaussian can be written as $\Sigma = R S S^T R^T$. To represent a scene, 3DGS optimizes 3D Gaussian positions $\boldsymbol{x}$, covariance $\Sigma$ (scaling $S$ and rotation $R$), opacity $\boldsymbol{o}$, and colours, represented by spherical harmonic (SH) coefficients, capturing view-dependent appearance. The optimization is interleaved with Gaussian adaptive density control (ADC). The model is optimized by rendering the learned Gaussians via a differentiable rasterizer, comparing the resulting image $I_r$ against the ground truth $I_{gt}$, and minimizing the loss function:

\begin{equation}
    \mathcal{L} = (1-\lambda)\mathcal{L}_1(I_r, I_{gt}) + \lambda \mathcal{L}_{\mathrm{SSIM}}(I_r,I_{gt}).
\end{equation}
\subsection{Sliding-Window Processing}
\label{sec:sliding_windows}

Using a single dynamic 3DGS representation to model an entire sequence becomes impractical for longer sequences, simply from a data processing standpoint. Furthermore, representing a lengthy sequence with a single model performs significantly worse than using multiple smaller fixed-sized segments, particularly if the scene has large motion or topological changes that cannot be modeled easily with one canonical representation and deformation field (Fig.~\ref{fig:performance_consistency}). To address this, we use a sliding-window approach, separating the sequence into smaller windows with overlapping frames. The window size is a hyperparameter whose effect is explored in Section~\ref{sec:ablation}. Each window comprises an independent dynamic 3DGS model, and we allow all 3D Gaussian parameters to change between windows, including the number of Gaussians, their positions, rotations, scaling, colours and opacities. The advantage of this approach is that independent models can be trained in parallel across multiple GPUs to speed up training.

\subsubsection{Adaptive Window Sampling}

We propose an adaptive method for choosing the appropriate sliding window sizes, balancing training overhead and model size with performance and temporal consistency. Given an input sequence of frame length $N_f$, we separate the sequence into smaller windows by adaptively sampling windows of different lengths $\{N_{w}\}$, depending on the amount of motion in the sequence. In high-motion areas, we aim to sample windows more frequently (shorter windows), and in low-motion areas, less frequently (longer windows). 

To do this, we leverage the magnitude of 2D optical flow from each camera viewpoint. We employ a greedy algorithm to adaptively select the sizes of windows, prior to training, based on the accumulated optical flow magnitude. Specifically, we estimate per-frame optical flow $\boldsymbol{f}$ using a pre-trained RAFT~\cite{Teed2020RAFTRA} model for each camera view $j \in V$ and all frames in the sequence $i \in N_f$, and compute the mean flow magnitude summed over each frame:

\begin{equation}
\hat{v}_i = \frac{1}{V}
    \sum^{V}_j 
    \sum^{N_f-1}_i
     || \boldsymbol{f} (I^j_i, I^j_{i+1} ) ||^2_2
\end{equation}

We iterate over each frame in the sequence with a greedy heuristic; spawning a new window when the sum of mean flow $\hat{v}_i$ exceeds a pre-defined threshold. This ensures that each window contains a similar amount of movement, leading to a balanced distribution of the total representational workload. Taking the average across viewpoints makes this approach somewhat invariant to the number of cameras, while placing a limit on the total flow stops an excessive amount of movement within each window. Note, each sampled window overlaps with the next window by a single frame, such that neighbouring windows share a common image frame. This is to enable inter-window temporal consistency (section~\ref{sec:temporal}).

\subsection{Dynamic 3D Gaussians}

To extend static 3DGS to handle dynamic scenes, we introduce a temporally-local dynamic MLP unique to each sliding window in the sequence (section~\ref{sec:sliding_windows}). Each time-dependent MLP learns a temporally-local deformation field, mapping from a learned per-window canonical space to a set of 3D Gaussians for each frame in the window. Each deformation field, represented by a small MLP $\mathcal{F}_{\theta}$ with weights $\theta$, takes as input the normalized frame time $t \in [0,1]$ and 3D Gaussian means $\boldsymbol{x}$ (normalized by the scene's mean and standard deviation), and outputs displacements to their positions $\Delta \boldsymbol{x}$, rotations $\Delta \boldsymbol{r}$ and scaling $\Delta \boldsymbol{s}$:

\begin{equation}
    \Delta \boldsymbol{x}(t), \Delta \boldsymbol{r}(t), \Delta \boldsymbol{s}(t)
    = \mathcal{F}_{\theta} (\gamma({\boldsymbol{x}}), \gamma(t))
\end{equation}
where $\gamma(.)$ denotes sinusoidal positional encoding $\gamma : \mathbb{R}^3 \rightarrow \mathbb{R}^{3+6m}$, $\gamma(x) = (x, \dots, \sin(2k \pi x), \cos(2k \pi x), \dots)$. We use small MLPs to reduce overfitting, setting the number of frequency components $m = 6$, MLP depth $D = 4$ and width $W = 16$, with two skip connections.

\subsection{Tunable Dynamic MLPs}
\label{sec:dynamic_mlp}

Ideally, time-dependent MLPs model motion in the scene by associating the dynamic parts with the temporal input component $\gamma(t)$ and thus learn to decouple the scene into i) static canonical 3D Gaussians and ii) dynamic 3D Gaussians. However, in scenes with an imbalance of static vs dynamic parts, \eg Neural 3D Video~\cite{li2022a} with mostly static backgrounds, the MLP struggles to disentangle static and dynamic regions, resulting in poorly estimated deformation fields.

To counteract this, we introduce tuneable dynamic MLPs~\cite{Maggioni2023TunableCW} (\ie MLPs with $M$ sets of weights) governed by $M$ sets of learnable blending parameters $\{ \boldsymbol{\alpha} \in \mathbb{R}^{M \times N_g} \}$. The tuning parameters weigh the respective parameters of the MLP for each input Gaussian $i \in N_g$, enabling the learning of $M$ modes of variation corresponding to different \textit{motion modes}. With $M=2$, the MLP has the ability to decouple static and dynamic Gaussians in a smoothly weighted manner. Given the set of blending parameters $\{ \alpha_i \}^M_{m=1}$ for the $i$-th input Gaussian $\boldsymbol{x}_i$, the output of a single layer of the dynamic MLP $\boldsymbol{y}_i$ can be written as a weighted sum over the $M$ sets of weights:

\begin{equation}
    \boldsymbol{y}_i = \phi \left( \sum^M_{m=1} \left( \alpha_{i,m} \boldsymbol{w}_m^T \boldsymbol{x}_i + \alpha_{i,m} \boldsymbol{b}_m \right) \right),
\end{equation}

\noindent where $\boldsymbol{w}$ and $\boldsymbol{b}$ are the weights and bias, and $\phi$ is a non-linear activation function. 
The blending parameters $\{ \alpha_i \}^M_{m=1}$ linearly blend $M$ sets of weights and biases of each layer of the MLP for each input Gaussian which, when passed through the activation function, enables nonlinear interaction between $\boldsymbol{\alpha}$ and the output. Applying $\boldsymbol{\alpha}$ in this way naturally enables the learning different of \textit{motion modes} with a single forward pass of the MLP. Thus the dynamic MLP $\mathcal{F}^{dyn}_{\theta}$ becomes a function of position, time, and blending parameters $\boldsymbol{\alpha}$:

\begin{equation}
    \Delta \boldsymbol{x}(t), \Delta \boldsymbol{r}(t), \Delta \boldsymbol{s}(t)
    = \mathcal{F}^{dyn}_{\theta} (\gamma({\boldsymbol{x}}), \gamma(t), \boldsymbol{\alpha}).
\end{equation}

We implement the dynamic MLP as a single batch matrix multiplication (see supplementary). Blending parameters $\boldsymbol{\alpha}$ are initialized as a binary mask of dynamic Gaussians (0-static, 1-dynamic) as follows. For a sliding window, we project all 3D Gaussians into each camera view: $u = \Pi_j ( {\boldsymbol{x}} )$, obtaining their 2D-pixel coordinates. We compute their L1 pixel differences from the central frame to each frame in the window. If the difference is larger than a threshold, we label that Gaussian 1, otherwise 0. To be robust to occlusions and mislabelling, we average the assigned label over all views and frames in the window. If the average is greater than 0.5, we initialize the Gaussian dynamic, otherwise static.

\begin{figure}
    \centering
    \begin{subfigure}[b]{0.48\textwidth}
         \centering
            \includegraphics[trim={0pt 0pt 750pt 0pt},clip,width=0.992\linewidth]{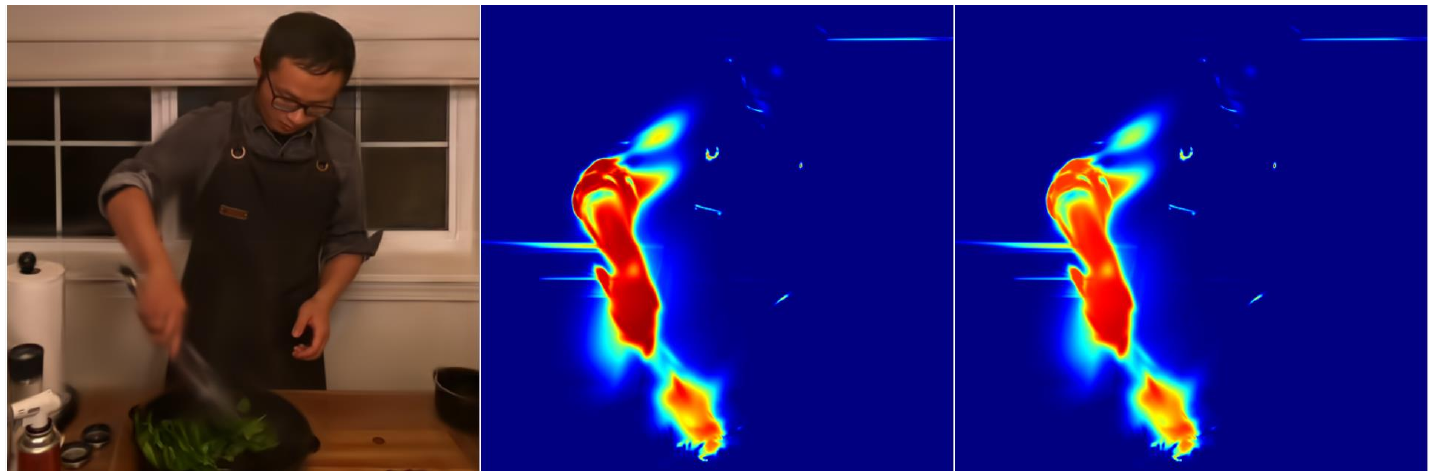}
         \label{fig:dyn_suba}
    \end{subfigure}
    \hfill
    \begin{subfigure}[b]{0.51\textwidth}
         \centering
         \includegraphics[trim={700pt 0pt 0pt 0pt},clip,width=1.0\linewidth]{images/SWAGS_dynamic_figure_small.pdf}
         \label{fig:dyn_subb}
     \end{subfigure}
    \caption{ 
    Dynamic MLPs with tunable parameters $\boldsymbol{\alpha}$ weigh the parameters of the MLP for each Gaussian. We show renders from two scenes, left:\,\textit{cook spinach}\cite{li2022a} and right:\,\textit{Train}\cite{Sabater2017}. Shown from left-to-right: image render, tunable $\boldsymbol{\alpha}$ parameters, and normalized MLP displacements $\Delta \boldsymbol{x}$. Note, $\boldsymbol{\alpha}$ highlights the scene's dynamic regions.}
    \label{fig:tuneableMLP}
\end{figure}

Once initialized, we set the blending parameter $\boldsymbol{\alpha}$ as a learnable parameter, and optimize it via back-propagation together with the rest of the system. We assume $\boldsymbol{\alpha}$ is constant for all frames in a sliding window to reduce the complexity of the optimization, which is a fair assumption given our adaptive window sampling mechanism. Fig.~\ref{fig:tuneableMLP} visualizes the learning of the MLP tuning parameter. We observe $\boldsymbol{\alpha}$ providing higher weight to Gaussians likely to be dynamic. Fig.~\ref{fig:tuneableMLP} (right) plots the magnitudes of resulting displacements $\Delta \boldsymbol{x}$ output by the MLP. 
Note, $\boldsymbol{\alpha}$ does not have to be entirely accurate, as the MLP learns to adjust accordingly, yet enables the MLP to handle highly imbalanced scenes (Table~\ref{table:ablation_technicolor2}).

\subsection{Temporal Consistency Fine-tuning}
\label{sec:temporal}
 
Due to the non-deterministic nature of the 3DGS optimization, independently trained models for each sliding window may produce slightly different results. When the resulting renders from each model are played back in sequence, noticeable flickering can be observed, particularly in novel views (though it's less noticeable in the training views). Consequently, after training models for all sliding windows separately, we tackle inter-window temporal flickering by introducing a fine-tuning step for temporal consistency. This step uses a self-supervising loss function to aid in smooth transitions between the models of separate windows.

We fine-tune each model for a short period (3000 iterations in our experiments). We initiate the process from the first sliding window and progress through the sequence sequentially. For a window comprising $N_w$ frames, we load the trained model and the model from the preceding window, with one overlapping frame between them. We then freeze all the parameters of the previous model. In order to fine-tune the model, as the flickering is mainly observed in novel views, we randomly sample novel test views in between the training views by rigidly interpolating the training camera poses $P_j = [R|\boldsymbol{t}] \in \mathbb{R}^{4 \times 4}$ in $\boldsymbol{\mathrm{SE}}(3)$:

\begin{equation}
    P_{\mathrm{novel}} = \exp_{\mathrm{M}} \sum^{V}_j \beta_j \log_{\mathrm{M}} ( P_j )
\end{equation}

\noindent where $\exp_\mathrm{M}$ and $\log_\mathrm{M}$ are the matrix exponential and logarithm~\cite{Alexa2002LinearCO} respectively, and $\beta_j \in [0,1]$ is a uniformly sampled weighting such that $\sum^{V}_j \beta_j = 1$.

In one fine-tuning step, we render the overlapping frames from a randomly sampled novel viewpoint using both the model of the current window $w$ and the previous window $w-1$. This means we use the first frame of current window $I^w_{t=0}$ and the last frame of the previous window $I^{w-1}_{t=N_w-1}$. We then apply a consistency loss, which is simply the L1 loss on the two image renders from both models:

\begin{equation}
    \mathcal{L}_{\mathrm{consistency}} = | I^w_{t=0} - I^{w-1}_{t=N_w-1} |_1
\end{equation}

When fine-tuning a model, we only allow the canonical (static) representation to change, while freezing the weights of the time-dependent dynamic MLP. Since the canonical set of 3D Gaussians is shared for all frames in a sliding window, we need to be careful not to negatively impact other frames when refining the overlapping frame. To address this, we use an alternating strategy for refinement, enforcing temporal consistency on the overlapping frame $75\%$ of the time and training with the remaining views and frames as usual for the remaining $25\%$.

\subsection{Implementation Details}

We implement our method in PyTorch, building upon the codebase and differentiable rasterizer of 3DGS~\cite{kerbl2023}. We initialize each model with a point cloud obtained from COLMAP~\cite{Schnberger2016StructurefromMotionR}. Each dynamic 3DGS model is trained for 15K iterations for all sampled windows in a sequence. The initial 2K iterations comprise a warm-up stage; training with only the central window frame and freezing the weights of the MLP, allowing the canonical representation to stabilize. We optimize the Gaussians' positions, rotations, scaling, opacities and SH coefficients. Afterwards, we unfreeze the MLP and allow the deformation field and tuning parameters $\boldsymbol{\alpha}$ to optimize. We find this staggered optimization approach leads to better convergence. The number of Gaussians densifies for 8K iterations, after which the number of Gaussians is fixed. Inputs to the MLP are normalized by the mean and standard deviation of the canonical point cloud post-warm-up stage before frequency encoding, leading to faster and stabler convergence. We train with Adam optimizer, using different learning rates for each parameter following the implementation of \cite{kerbl2023}. The learning rate for the MLP and $\boldsymbol{\alpha}$ parameters are set to 1e-4, with the MLP learning rate undergoing exponential decay by factor 1e-2 in 20K iterations. During the initial optimization phase, due to the independence of each Gaussian model, we train each model in parallel on eight 32Gb Tesla V100 GPUs to speed up training. Afterwards, we perform temporal fine-tuning of each model sequentially using a single GPU for 3K iterations each.

%% file: sec/4_results.tex
\begin{table}[t]
\centering
\renewcommand{\arraystretch}{1.0}
\caption{Quantitative results on the Neural 3D Video dataset~\cite{li2022a}, averaged over all scenes. \colorbox{red!30}{Best} and \colorbox{orange!30}{second best} highlighted. Our method performs best overall whilst enabling real-time frame rates. Per-scene breakdown of results are given in Table~\ref{table:table2}.}
\renewcommand{\arraystretch}{0.8}
\begin{tabular}{l c c c c} \toprule
    Method & PSNR & SSIM & LPIPS & FPS\\ \midrule
    MixVoxels~\cite{wang2023}  & 30.42 & 0.923 & 0.124 & {4.30}\\
    K-Planes~\cite{fridovich2023}  & 30.63 & 0.922 & 0.117 & 0.33 \\
    HexPlane~\cite{cao2023} & 30.00 & 0.922 & 0.113 & 0.24 \\
    HyperReel~\cite{attal2023} & \colorbox{orange!30}{30.78} & \colorbox{orange!30}{0.931} & \colorbox{orange!30}{0.101} & 3.60 \\
    NeRFPlayer~\cite{song2023}  & 30.70 & \colorbox{orange!30}{0.931} & 0.121 & 0.10 \\
    StreamRF~\cite{li2022b} & 30.23 & 0.904 & 0.177 & 9.40 \\
    4DGS~\cite{Wu_2024_CVPR} & 27.61 & 0.916 & 0.135 & \colorbox{orange!30}{30.00} \\
    \textbf{Ours} & \colorbox{red!30}{31.10} & \colorbox{red!30}{0.940} & \colorbox{red!30}{0.096} & \colorbox{red!30}{71.51} \\ 
    \bottomrule
\end{tabular}
\label{table:table1}
\end{table}

\begin{figure}
  \begin{minipage}[b]{.54\linewidth}
    \centering
    \renewcommand{\arraystretch}{1.0}
    \captionof{table}{Quantitative results on Technicolor dataset~\cite{Sabater2017} at full resolution. \colorbox{red!30}{Best} and \colorbox{orange!30}{second best} results are highlighted.}
    \begin{tabular}{l c c c c} \toprule
        Method & PSNR & SSIM & LPIPS & FPS\\ \midrule
        DyNeRF~\cite{li2022a} & 31.80 & 0.911 & 0.142 & 0.02 \\
        HypeRreel~\cite{attal2023} & \colorbox{orange!30}{32.50}  & 0.902  & \colorbox{red!30}{0.113} & 0.45  \\
        Dynamic3DG~\cite{luiten2024} & 27.02 & 0.832 & 0.228 & \colorbox{red!30}{86.96} \\
        \textbf{Ours} & \colorbox{red!30}{33.65} & \colorbox{red!30}{0.934} & \colorbox{orange!30}{0.117} & \colorbox{orange!30}{23.79} \\ 
        \bottomrule
    \end{tabular}
    \label{table:technicolor_avg}
  \end{minipage}\hfill
  \begin{minipage}[b]{.42\linewidth}
    \centering
    \includegraphics[trim={40pt 0 70pt 40pt},clip,width=0.92\linewidth]{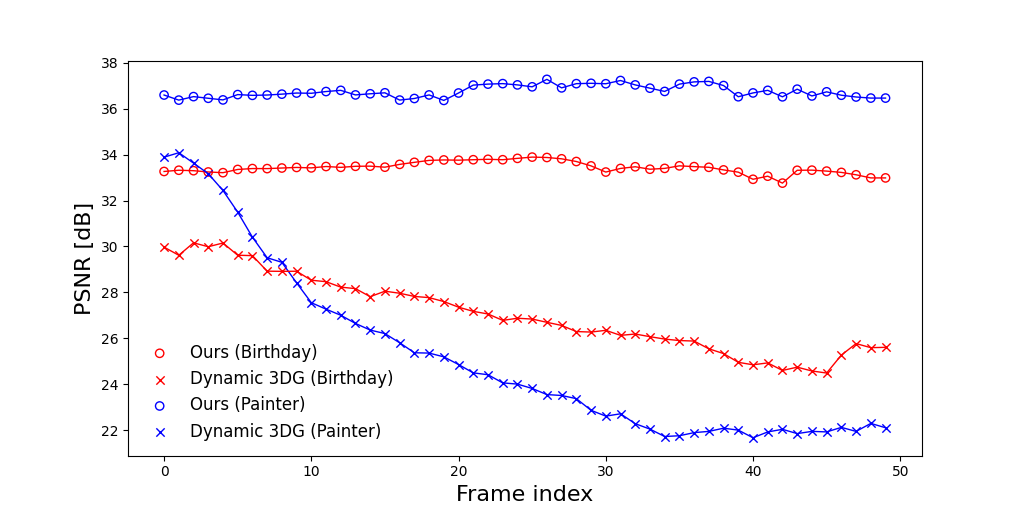}
    \captionof{figure}{Comparison in performance consistency for \textbf{Ours} and Dynamic3DG in consecutive frames.} 
    \label{fig:performance_consistency}
  \end{minipage}
\end{figure}

\begin{table}
\centering
\caption{Quantitative results on the Neural 3D Video dataset~\cite{li2022a}, evaluated for 300 frames at $1352 \times 1014$ resolution. \textdagger As reported in \cite{attal2023}, \textdaggerdbl Natively trained in lower resolution, upscaled. \colorbox{red!30}{Best} and \colorbox{orange!30}{second best} results highlighted respectively. Our method performs competitively in all metrics, usually coming in either first or second place.}
\renewcommand{\arraystretch}{0.8}
\begin{tabularx}{\linewidth}{lYYYYYYYYY}
\toprule
Scene & \multicolumn{3}{c}{Cook Spinach} & \multicolumn{3}{c}{Cut Roast Beef} & \multicolumn{3}{c}{Flame Steak} \\
\cmidrule(lr){2-4}\cmidrule(lr){5-7}\cmidrule(lr){8-10}
    Method & PSNR & SSIM & LPIPS & PSNR & SSIM & LPIPS & PSNR & SSIM & LPIPS \\
    \midrule
    MixVoxels~\cite{wang2023} & $31.39$ & $0.931$ & $0.113$ & $31.38$ & $0.928$ & $0.111$ & $30.15$ & $0.938$ & $0.108$ \\
    K-Planes~\cite{fridovich2023} & $31.23$ & $0.926$ & $0.114$ & \colorbox{orange!30}{$31.87$} & $0.928$ & $0.114$ & $31.49$ & $0.940$ & $0.102$  \\
    HexPlane\textsuperscript{\textdaggerdbl}~\cite{cao2023} & $31.05$ & $0.928$ & $0.114$ & $30.83$ & $0.927$ & $0.115$ & $30.42$ & $0.939$ & $0.104$  \\ 
    HyperReel~\cite{attal2023} & \colorbox{orange!30}{$31.77$} & \colorbox{orange!30}{$0.932$} & \colorbox{red!30}{$0.090$} & \colorbox{red!30}{$32.25$} & \colorbox{orange!30}{$0.936$} & \colorbox{red!30}{$0.086$} & $31.48$ & $0.939$ & \colorbox{red!30}{$0.083$} \\
    NeRFPlayer\textsuperscript{\textdagger}~\cite{song2023} & $30.58$ & $0.929$ & $0.113$ & $29.35$ & $0.908$ & $0.144$ & \colorbox{orange!30}{$31.93$} & \colorbox{orange!30}{$0.950$} & $0.088$ \\ 
    StreamRF~\cite{li2022b} & $30.89$ & $0.914$ & $0.162$ & $30.75$ & $0.917$ & $0.154$ & $31.37$ & $0.923$ & $0.152$  \\ 
    \textbf{Ours}  & \colorbox{red!30}{31.96} & \colorbox{red!30}{0.946} & \colorbox{orange!30}{0.094} & 31.84 & \colorbox{red!30}{0.945} & \colorbox{orange!30}{0.099} & \colorbox{red!30}{32.18} & \colorbox{red!30}{0.953} & \colorbox{orange!30}{0.087}  \\ 
    \midrule
    & \multicolumn{3}{c}{Sear steak} & \multicolumn{3}{c}{Coffee Martini} & \multicolumn{3}{c}{Flame Salmon} \\
    \midrule
    MixVoxels~\cite{wang2023} & $30.85$ & $0.940$ & $0.103$ & $29.25$ & $0.901$ & $0.147$ & $29.50$ & $0.898$ & $0.163$ \\
    K-Planes~\cite{fridovich2023} & $30.28$ & $0.937$ & $0.104$ & \colorbox{orange!30}{$29.30$} & $0.900$ & $0.134$ & \colorbox{orange!30}{$29.58$} & $0.901$ & $0.132$ \\ 
    HexPlane\textsuperscript{\textdaggerdbl}~\cite{cao2023}  & $30.00$ & $0.939$ & $0.105$ & $28.45$ & $0.891$ & $0.149$ & $29.23$ & $0.905$ & \colorbox{red!30}{$0.088$} \\ 
    HyperReel~\cite{attal2023} & \colorbox{orange!30}{$31.88$} & \colorbox{orange!30}{$0.942$} & \colorbox{red!30}{$0.080$} & $28.65$ & $0.897$ & $0.129$ & $28.26$ & \colorbox{red!30}{$0.941$} & $0.136$\\
    NeRFPlayer\textsuperscript{\textdagger}~\cite{song2023} & $29.13$ & $0.908$ & $0.138$ & \colorbox{red!30}{$31.53$} & \colorbox{red!30}{$0.951$} & \colorbox{red!30}{$0.085$} & \colorbox{red!30}{$31.65$} & \colorbox{orange!30}{$0.940$} & \colorbox{orange!30}{$0.098$}\\
    StreamRF~\cite{li2022b} & $31.60$ & $0.925$ & $0.147$ & $28.13$ & $0.873$ & $0.219$ & 28.69  & 0.872 & 0.229\\
    \textbf{Ours} & \colorbox{red!30}{32.21} & \colorbox{red!30}{0.950} & \colorbox{orange!30}{0.092} & 29.16 & \colorbox{orange!30}{0.921} & \colorbox{orange!30}{0.105} & 29.25 & 0.925 & 0.100\\ 
    \bottomrule
\end{tabularx}
\label{table:table2}
\end{table}

\begin{table}
\centering
\caption{Quantitative results on the Technicolor dataset~\cite{Sabater2017} evaluated at full resolution. \colorbox{red!30}{Best} and \colorbox{orange!30}{second best} results highlighted respectively.}
\renewcommand{\arraystretch}{0.8}
\begin{tabularx}{\linewidth}{lYYYYYYYYY} \toprule
    Scene & \multicolumn{3}{c}{Birthday} & \multicolumn{3}{c}{Fabien} & \multicolumn{3}{c}{Painter} \\
    \cmidrule(lr){2-4} \cmidrule(lr){5-7} \cmidrule(lr){8-10}
    Method & PSNR & SSIM & LPIPS & PSNR & SSIM & LPIPS & PSNR & SSIM & LPIPS\\
    \midrule
    DyNeRF~\cite{li2022a} & 29.20 & 0.909 & 0.067 & \colorbox{orange!30}{32.76} & \colorbox{orange!30}{0.909} & 0.242 & \colorbox{orange!30}{35.95} & \colorbox{orange!30}{0.930} & 0.147  \\
    HyperReel~\cite{attal2023} & \colorbox{orange!30}{30.79} & \colorbox{orange!30}{0.922} & \colorbox{orange!30}{0.062} & 32.28 & 0.860 & \colorbox{orange!30}{0.217} & 35.68 & 0.926 & \colorbox{red!30}{0.123} \\
    Dynamic3DG~\cite{luiten2024} & 27.06 & 0.859 & 0.113 & 26.34 & 0.834 & 0.268 & 25.18 & 0.758 & 0.395 \\
    
    \textbf{Ours} & \colorbox{red!30}{33.44} & \colorbox{red!30}{0.959} & \colorbox{red!30}{0.042} & \colorbox{red!30}{34.43} & \colorbox{red!30}{0.925} & \colorbox{red!30}{0.171} & \colorbox{red!30}{36.76} & \colorbox{red!30}{0.948} & \colorbox{orange!30}{0.128}  \\ 
    \midrule
    Scene & \multicolumn{3}{c}{Theater} & \multicolumn{3}{c}{Train} & \multicolumn{3}{c}{Average} \\
    \cmidrule(lr){2-4} \cmidrule(lr){5-7} \cmidrule(lr){8-10}
    DyNeRF~\cite{li2022a} & 29.53 & 0.875 & \colorbox{orange!30}{0.188} & \colorbox{orange!30}{31.58} & \colorbox{orange!30}{0.933} & 0.067 & 31.80 & \colorbox{orange!30}{0.911} & 0.142\\
    HyperReel~\cite{attal2023} &  \colorbox{red!30}{33.67} & \colorbox{red!30}{0.895} & \colorbox{red!30}{0.104} & 30.10 & 0.909 & \colorbox{orange!30}{0.061} & \colorbox{orange!30}{32.50} & 0.902 & \colorbox{orange!30}{0.113}\\
    Dynamic3DG~\cite{luiten2024} & 28.05 & 0.799 & 0.277 & 28.46 & 0.908 & 0.088 & 27.02 & 0.832 & 0.228\\
    \textbf{Ours} & \colorbox{orange!30}{29.81} & \colorbox{orange!30}{0.884} & 0.201  & \colorbox{red!30}{33.99} & \colorbox{red!30}{0.957} & \colorbox{red!30}{0.043} & \colorbox{red!30}{33.69} & \colorbox{red!30}{0.934} & \colorbox{orange!30}{0.117} \\ 
    \bottomrule
\end{tabularx}
\label{table:technicolor}
\end{table}

\section{Results}
\label{sec:results}
We evaluate our method on two real-world multi-view dynamic benchmarks: the Neural 3D Video dataset~\cite{li2022a} and the Technicolor dataset~\cite{Sabater2017}.

\begin{figure}
\captionsetup[subfigure]{labelformat=empty}
\centering
\begin{tabularx}{0.9\linewidth}{
l @{\hspace{0.85\tabcolsep}} 
c @{\hspace{0.85\tabcolsep}}
c @{\hspace{0.85\tabcolsep}}
c @{\hspace{0.85\tabcolsep}}
r}
\subfloat[MixVoxels]{\includegraphics[trim = {0, 374pt, 0, 0}, clip, width = 0.17\linewidth]{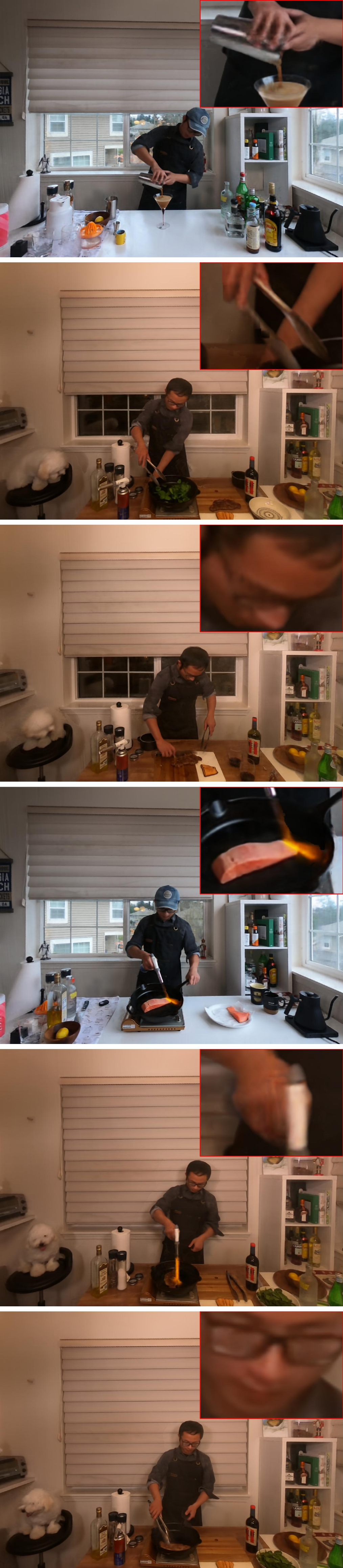}} &
\subfloat[K-Planes]{\includegraphics[trim = {0, 374pt, 0, 0}, clip, width = 0.17\linewidth]{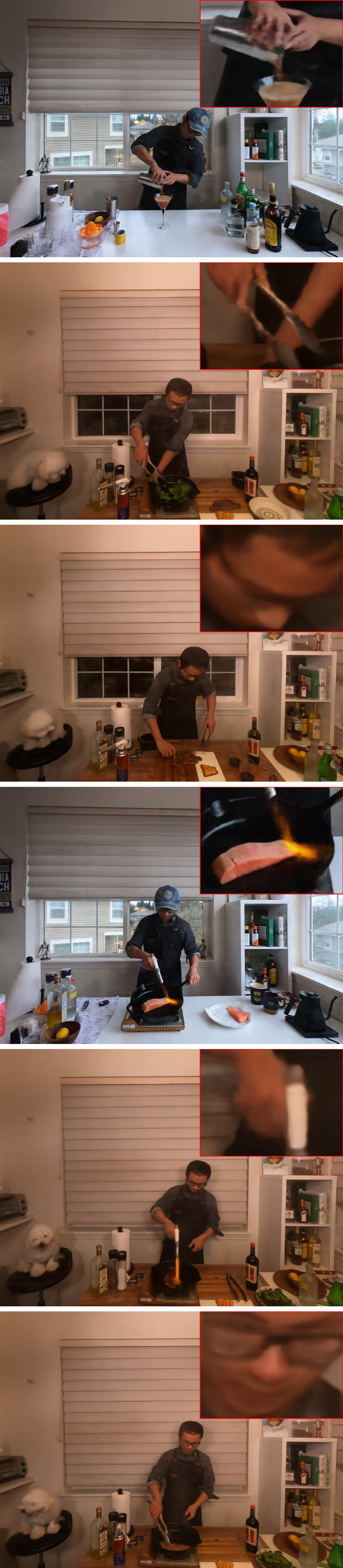}} &
\subfloat[HyperReel]{\includegraphics[trim = {0, 374pt, 0, 0}, clip, width = 0.17\linewidth]{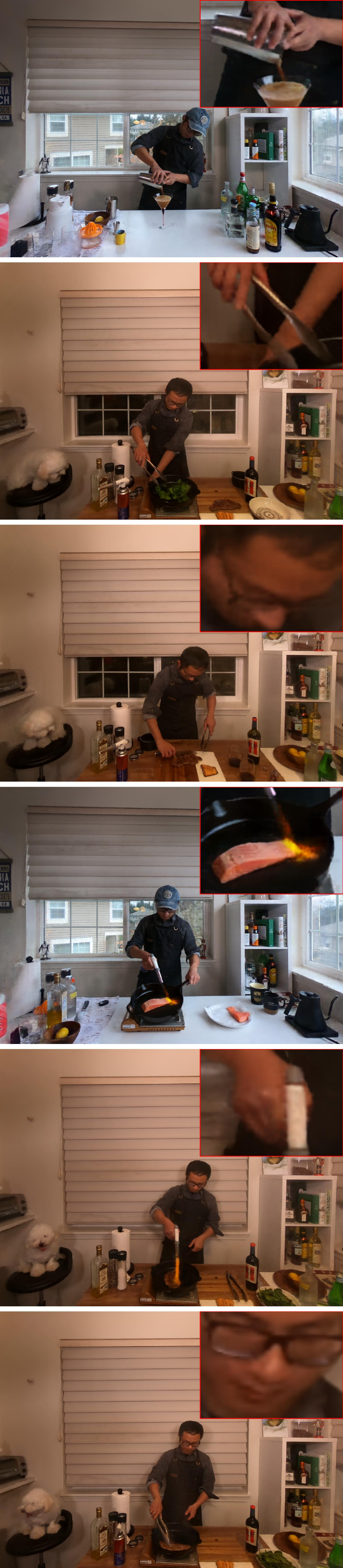}} &
\subfloat[\textbf{Ours}]{\includegraphics[trim = {0, 374pt, 0, 0}, clip, width = 0.17\linewidth]{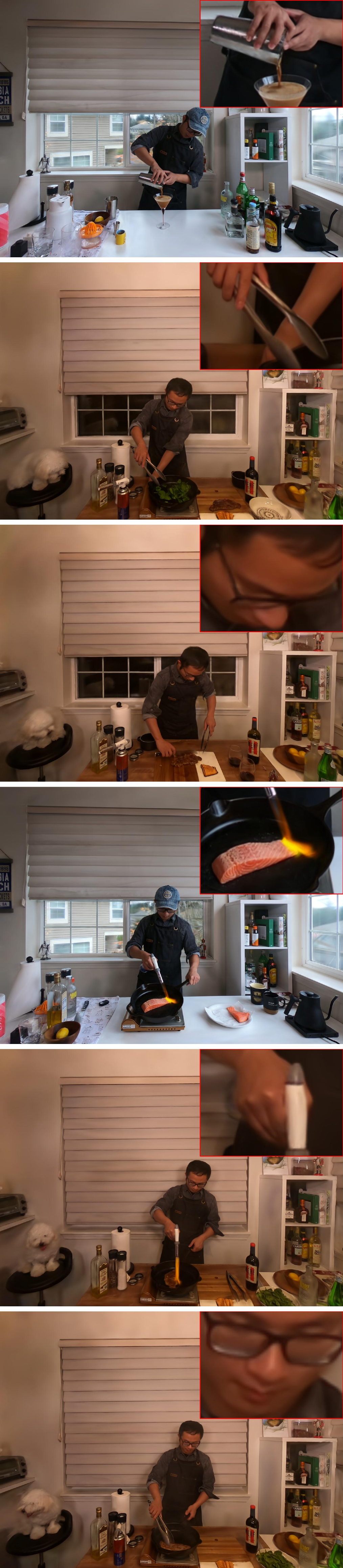}} &
\subfloat[Ground truth]{\includegraphics[trim = {0, 374pt, 0, 0}, clip, width = 0.17\linewidth]{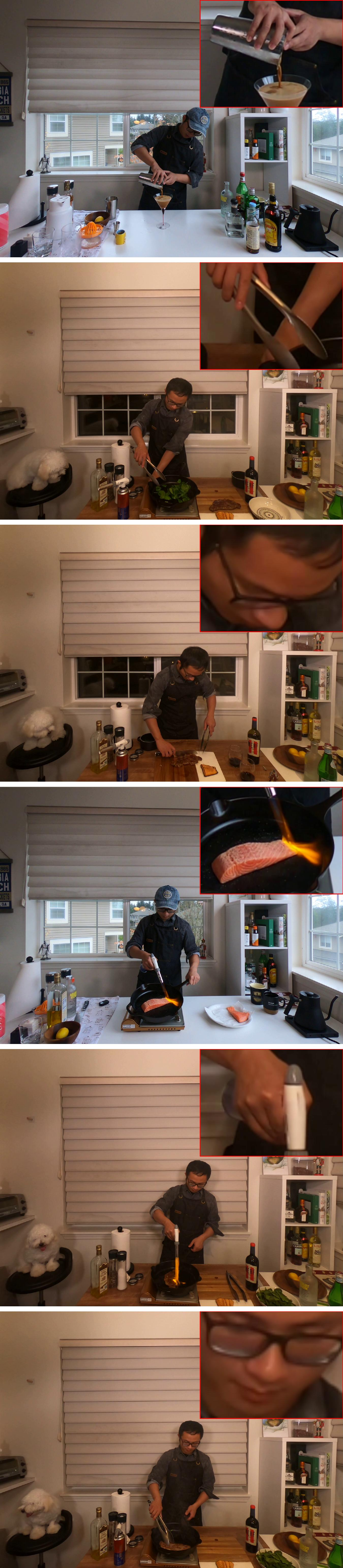}}
\end{tabularx}
\caption{Qualitative results on Neural 3D Video~\cite{li2022a}. Scenes top to bottom: i) \textit{coffee martini}, ii) \textit{cook spinach}, iii) \textit{cut roasted beef}, iv) \textit{flame salmon}, v) \textit{flame steak}.}
\label{fig:n3d_results}
\end{figure}

\noindent
\textbf{Neural 3D Video} comprises real-world dynamic scenes captured with a time-synchronized multi-view system at $2028 \times 2704$ resolution at 30 FPS. Camera parameters are estimated using COLMAP~\cite{Schnberger2016StructurefromMotionR}. We compare our method to K-Planes~\cite{fridovich2023}, HexPlane~\cite{cao2023}, MixVoxels~\cite{wang2023}, HyperReel~\cite{attal2023}, NeRFPlayer~\cite{song2023}, StreamRF~\cite{li2022b}, and 4DGS~\cite{Wu_2024_CVPR}. We compute quantitative metrics for the central test view at half the original resolution ($1014 \times 1352$) for 300 frames. The average results for each method over the dataset are given in Table~\ref{table:table1}, while Table~\ref{table:table2} provides a breakdown of the per-scene performance. The results show that our method performs best regarding PSNR and SSIM metrics while offering the fastest rendering performance. Qualitative results are shown in Fig.\,\ref{fig:n3d_results}.

\begin{figure}
\captionsetup[subfigure]{labelformat=empty}
\centering
\begin{tabularx}{0.88\linewidth}{
l @{\hspace{0.85\tabcolsep}} 
c @{\hspace{0.85\tabcolsep}}
c @{\hspace{0.85\tabcolsep}}
r}
\subfloat[HyperReel]{\includegraphics[width = 0.21\linewidth]{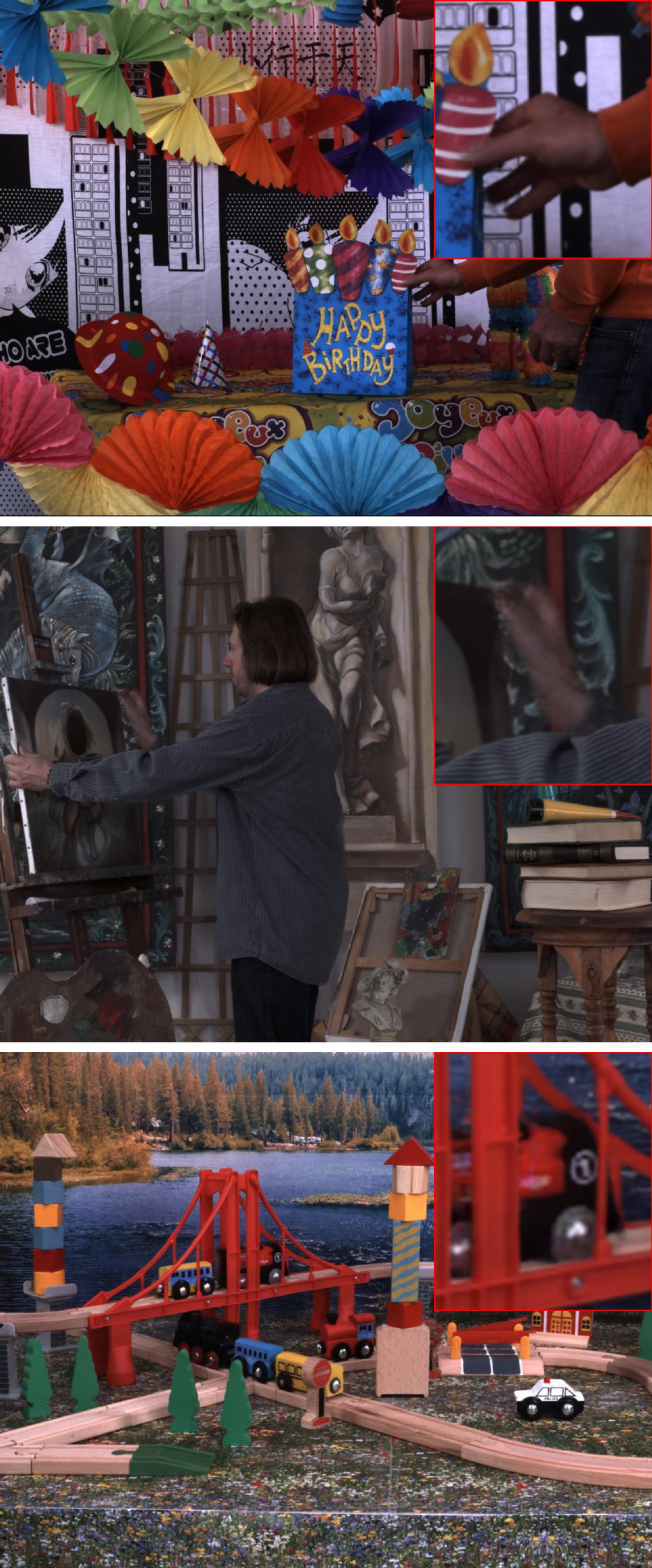}} &
\subfloat[Dynamic3DG]{\includegraphics[width = 0.21\linewidth]{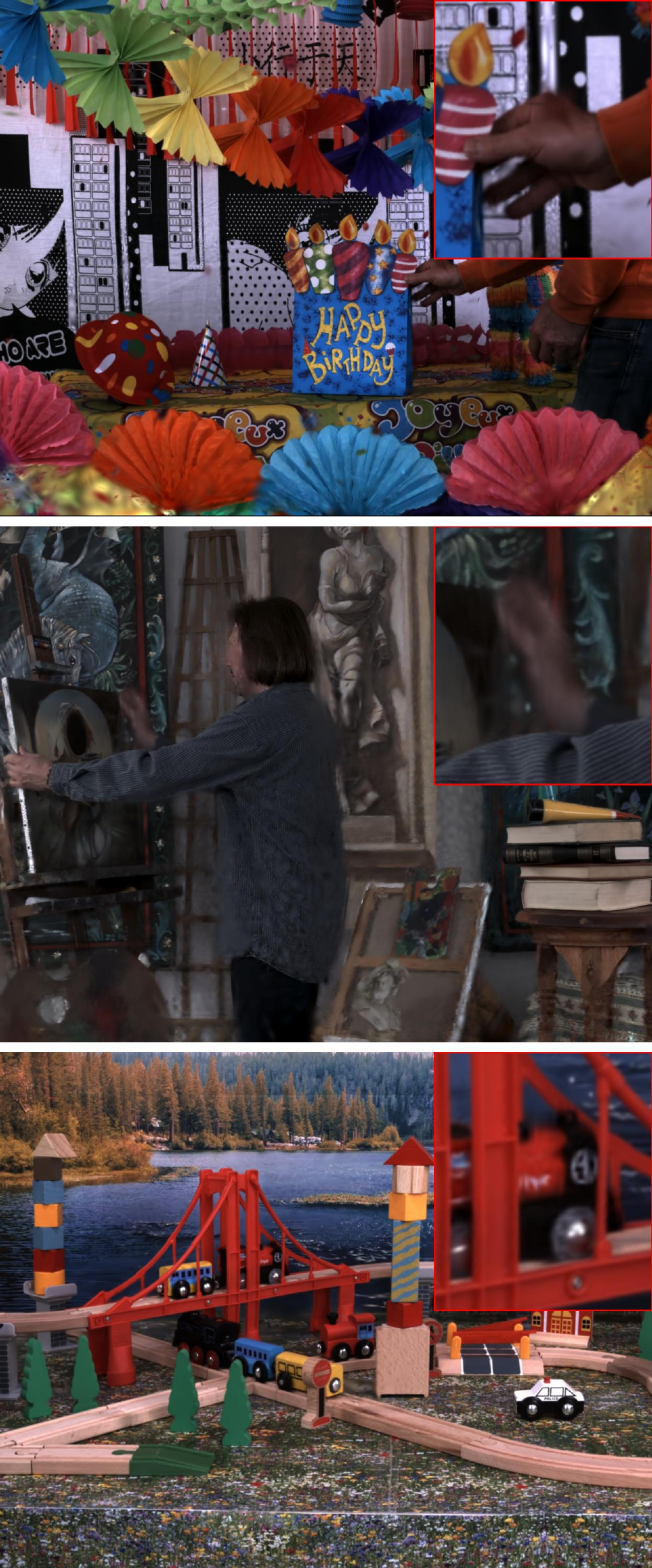}} &
\subfloat[\textbf{Ours}]{\includegraphics[width = 0.21\linewidth]{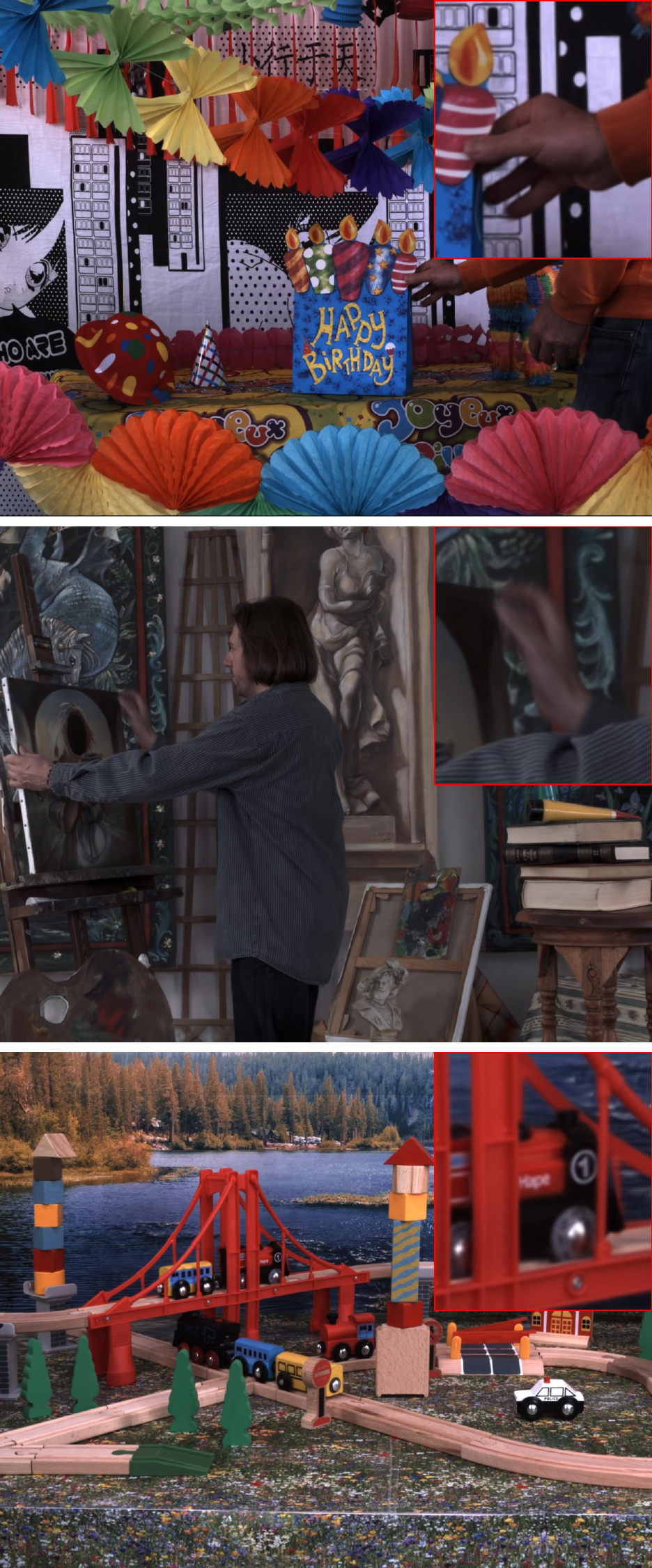}} &
\subfloat[Ground truth]{\includegraphics[width = 0.21\linewidth]{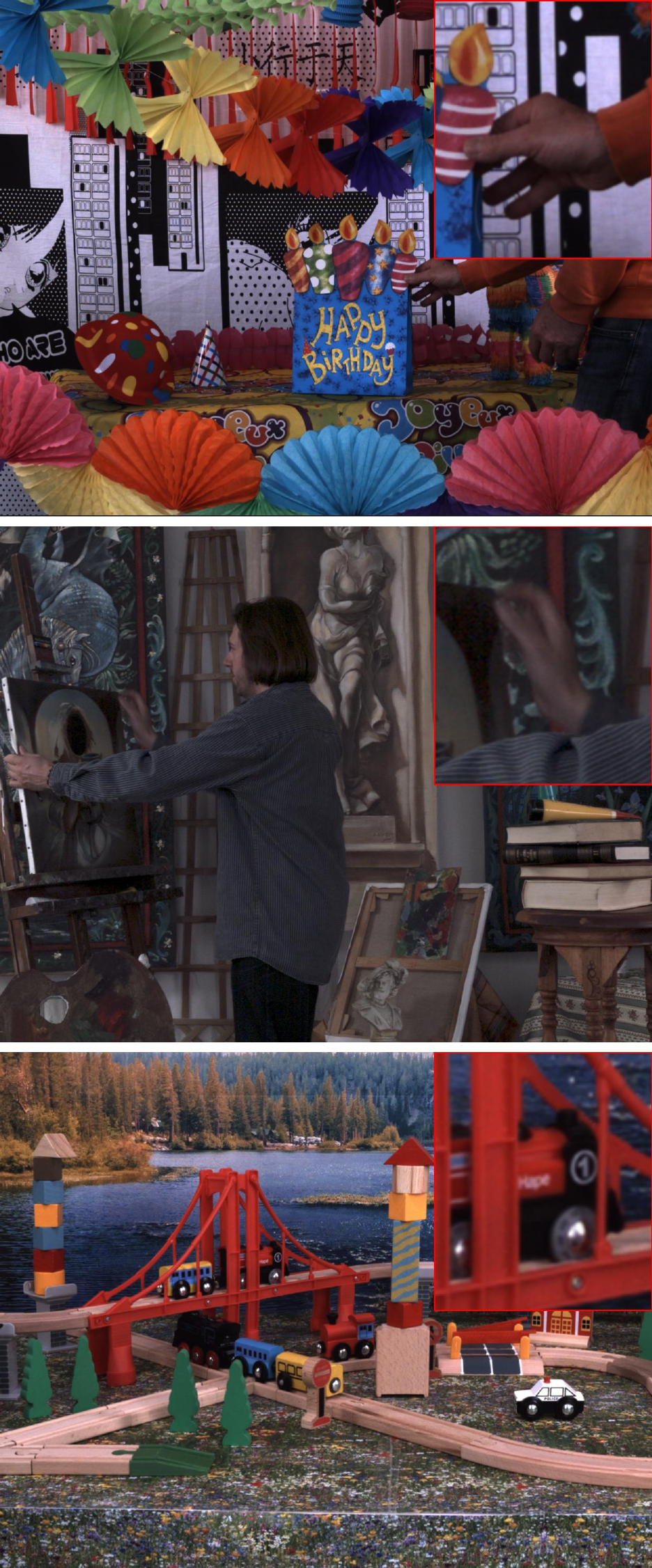}}
\end{tabularx}
\caption{Qualitative results on Technicolor~\cite{Sabater2017}. Scenes top to bottom: i) \textit{Birthday}, ii) \textit{Painter}, iii) \textit{Train}.}
\label{fig:technicolor_results}
\end{figure}

\noindent
\textbf{Technicolor} captures real dynamic scenes from a synchronized $4 \times 4$ camera array at $2048 \times 1088$ resolution. Following~\cite{attal2023}, we evaluate on the second row second column camera on five scenes: \textit{Birthday}, \textit{Fabien}, \textit{Painter}, \textit{Theater} and \textit{Trains}. We compare to DynNeRF~\cite{li2022a}, HyperReel~\cite{attal2023} and Dynamic 3D Gaussians~\cite{luiten2024}, with quantitative and qualitative results in Table~\ref{table:technicolor} and Fig.\,\ref{fig:technicolor_results} respectively.

\subsection{Ablation Studies}
\label{sec:ablation}

\begin{figure}[]
\centering
  \includegraphics[width=1.0\textwidth]{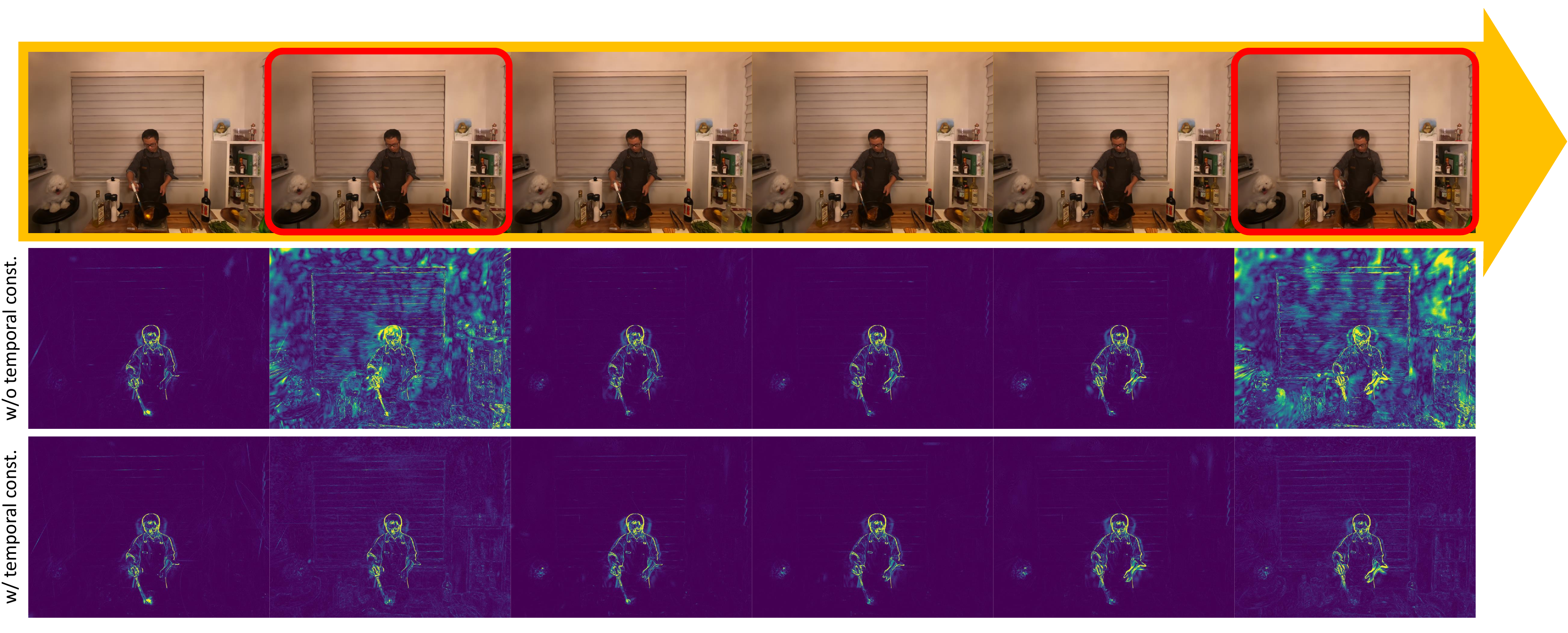}
  \caption{Ablation on temporal fine-tuning: we display the absolute error between neighbouring frame renders, with overlapping frames highlighted red. After fine-tuning, the error is substantially reduced and overall perceptual video quality is improved.}
  \label{fig:ablation}
\end{figure}

This section provides ablations showing the effectiveness of our sliding window and self-supervised temporal consistency fine-tuning strategies. As we train independent models for each window, flickering artifacts can occur in the final renders. To visualize this, Fig.~\ref{fig:ablation} plots absolute image error between renders of neighbouring frames (overlapping frames outlined in red). Without temporal consistency, we observe a spike in absolute error on overlapping frames, resulting in undesirable flickering. However, after fine-tuning, the error in overlapping frames is drastically reduced. In Table~\ref{table:table3}, we compute per-frame image metrics and estimate a measure of the temporal consistency using SoTA video quality assessment metric FAST-VQA~\cite{Wu2022FASTVQAEE}, where the quality score is in the range [0,1]. The table provides results averaged over all scenes from the Neural 3D Video dataset~\cite{li2022a}. Although we incur a minor penalty in some per-frame performance metrics (SSIM and LPIPS), we obtain a significantly higher video quality (VQA) score. This indicates a substantial improvement in temporal consistency and overall perceptual video quality, resulting in more pleasing renderings. 

\begin{table}
\centering
\renewcommand{\arraystretch}{0.8}
\caption{Ablation on temporal fine-tuning. Results averaged over all scenes from~\cite{li2022a}. Temporal consistency is measured using t-LPIPS~\cite{Chu2020LearningTC} and FAST-VQA~\cite{Wu2022FASTVQAEE} (quality score in the range [0,1]). Per-frame performance remains fairly constant, but temporal consistency and overall perceptual video quality is significantly improved.}
\renewcommand{\arraystretch}{0.8}
\begin{tabular}{
l @{\hspace{5\tabcolsep}}
c @{\hspace{5\tabcolsep}} 
c @{\hspace{5\tabcolsep}} 
c @{\hspace{5\tabcolsep}}
c @{\hspace{5\tabcolsep}}
c} \toprule
    Method  & PSNR\,$\uparrow$ & SSIM\,$\uparrow$ & LPIPS\,$\downarrow$ & t-LPIPS\,$\downarrow$ & VQA\,$\uparrow$ \\ \midrule
    w/o temporal consistency & 32.01 & \textbf{0.956} & \textbf{0.085} & \textbf{0.0129} & 0.666 \\
    w/ temporal consistency & \textbf{32.05} & 0.949 & 0.093 & \textbf{0.0102} & \textbf{0.726}\\
    Ground truth & - & - & - & - & 0.763\\
    \bottomrule
\end{tabular}
\label{table:table3}
\end{table}

\begin{table}
\centering
\caption{Ablation on sliding window size vs adaptive on \textit{Birthday} scene~\cite{Sabater2017}. We show the improvement from the dynamic MLP. Adaptive chooses window sizes that match, and in some metrics exceed, the best fixed-size window performance, striking balance between performance, temporal consistency (t-LPIPS) and training time (GPU hrs).\label{table:ablation_technicolor2}}
\renewcommand{\arraystretch}{0.8}
\begin{tabularx}{\linewidth}{XYYcccc} 
    & & & & & & \\
    \toprule
    {\footnotesize Window Size
    } & {\footnotesize No. Windows} & {\footnotesize Train time
    } & {\footnotesize PSNR} & {\footnotesize SSIM} & {\footnotesize LPIPS} & {\footnotesize t-LPIPS} \\ \midrule
    3 & 24 & 16.0 & 33.12 & 0.956 & 0.048 & 0.0076 \\
    9 & 6 & 4.0 & 33.38 & 0.959 & 0.043 & 0.0053 \\ 
    17 & 3 & 2.0 & 33.01 & 0.956 & 0.045 & 0.0049\\ 
    25 & 2 & 1.3 & 32.97 & 0.956 & 0.043 & 0.0048\\ 
    49 & 1 & 0.7 & 32.73 & 0.955 & 0.047 & 0.0051\\
    \textbf{Adaptive} & 5 & 3.3 & 33.44 & 0.959 & 0.042 & 0.0051  \\ 
    {\footnotesize Adaptive\,(w/o~dyn.~MLP)} & 5 & 3.3 & 32.76 & 0.957 & 0.045 & 0.0062 \\
    \bottomrule
\end{tabularx}
\end{table}

\begin{figure}[h!]
\centering
  \includegraphics[width=1.0\textwidth]{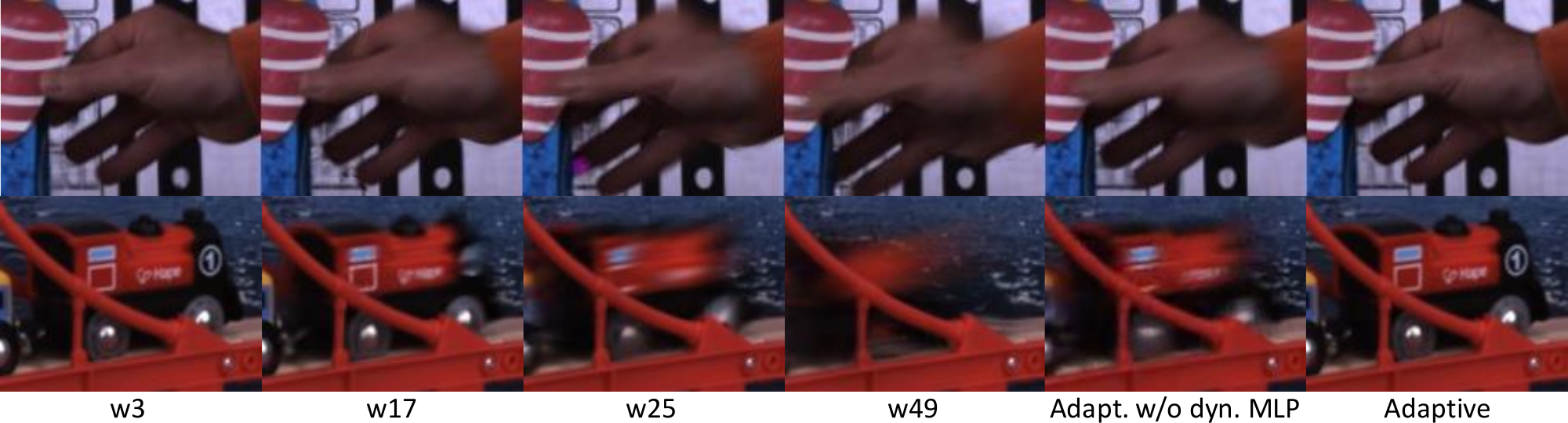}
  \caption{Ablation on sliding window size vs adaptive window sampling (with and without dynamic MLP) on scenes from the Technicolor dataset~\cite{Sabater2017}. The single-window scenario (w49) has the shortest training time but unsatisfactory visual quality, while adaptive windows offer the best balance between quality and training overhead.}
  \label{fig:ablation_vis}
\end{figure}

Table~\ref{table:ablation_technicolor2} and Fig.~\ref{fig:ablation_vis} show the impact of the sliding-window size hyperparameter and our adaptive sampling strategy. Adaptive sampling automatically chooses appropriate window sizes that match and sometimes exceed the best fixed-size window performance, striking a good balance between performance, temporal consistency (t-LPIPS~\cite{Chu2020LearningTC}) and training time. Adaptive performs the best overall with fewer windows (less storage requirements). We also show the advantage gained from using a dynamic MLP vs a regular MLP, which improves all metrics.

%% file: sec/5_conclusion.tex
\section{Conclusion}
\label{sec:conclusion}

We have presented a method to render novel views of dynamic scenes by extending the 3DGS framework. Results show that our method produces high-quality renderings, even with complex motions \eg flames. Key to our approach is sliding-window processing, which adaptively partitions sequences into manageable chunks. Processing each window separately, allowing the canonical representation and deformation field to vary throughout the sequence, enables us to handle complex topological changes, and reduces the magnitude and variability of the 3D scene flow. In contrast, other methods that learn a single representation for whole sequences are impractical for long sequences and degrade in quality with increasing length. Introducing an MLP for each window learns the deformation field from each canonical representation to a set of per-frame 3D Gaussians. Moreover, learnable tuning parameters help disentangle static and dynamic parts of the scene, which we find essential for imbalanced scenes. Our ablations show self-supervised temporal consistency fine-tuning reduces temporal flickering and improves the overall perceptual video quality, with only a minor impact on per-frame performance metrics. Overall, our method performs strongly compared to recent SoTA quantitatively and obtains sharper, temporally-consistent results.